\relax
\documentclass[letterpaper]{article} 
\usepackage{aaai22}
\usepackage{amsmath}
\usepackage{times}  
\usepackage{helvet}  
\usepackage{courier}  
\usepackage[hyphens]{url}  
\usepackage{graphicx} 
\usepackage{natbib}  
\usepackage{caption} 
\usepackage{algorithm}
\usepackage{algorithmic}

\usepackage{newfloat}
\usepackage{listings}
\DeclareCaptionStyle{ruled}{labelfont=normalfont,labelsep=colon,strut=off} 
\floatstyle{ruled}
\newfloat{listing}{tb}{lst}{}
\floatname{listing}{Listing}
\title{Forecast-Aware Model Driven LSTM}
\author {
    Sophia Hamer,\textsuperscript{\rm 1}
    Jennifer Sleeman,\textsuperscript{\rm 1}
    Ivanka Stajner\textsuperscript{\rm 2}
}
\affiliations {
    \textsuperscript{\rm 1} University of Maryland, Baltimore County\\
    Department of Computer Science and Electrical Engineering \\ 
    1000 Hilltop Cir, Baltimore, MD 21250\\
    \textsuperscript{\rm 2} National Oceanic and Atmospheric Administration\\
    5830 University Research Ct, College Park, MD 20740\\
   \{chamer1, jsleem1\}@umbc.edu, 
    ivanka.stajner@noaa.gov
}

\begin{document}

\maketitle

\begin{abstract}
Poor air quality can have a significant impact on human health.  The National Oceanic and Atmospheric Administration (NOAA) air quality forecasting guidance is challenged by the increasing presence of extreme air quality events due to extreme weather events such as wild fires and heatwaves. These extreme air quality events further affect human health.  Traditional methods used to correct model bias make assumptions about linearity and the underlying distribution.  Extreme air quality events tend to occur without a strong signal leading up to the event and this behavior tends to cause existing methods to either under or over compensate for the bias.  Deep learning holds promise for air quality forecasting in the presence of extreme air quality events due to its ability to generalize and learn nonlinear problems.  However, in the presence of these anomalous air quality events, standard deep network approaches that use a single network for generalizing to future forecasts, may not always provide the best performance even with a full feature-set including geography and meteorology.  In this work we describe a method that combines unsupervised learning and a forecast-aware bi-directional Long Short-Term Memory (LSTM) network to perform bias correction for operational air quality forecasting using AirNow station data for ozone and PM\textsubscript{2.5} in the continental US.  Using an unsupervised clustering method trained on station geographical features such as latitude and longitude, urbanization, and elevation, the learned clusters direct training by partitioning the training data for the LSTM networks.  LSTMs are forecast-aware and implemented using a unique way to perform learning forward and backwards in time across forecasting days.  When comparing the Root Mean Squared Error (RMSE) of the forecast model to the RMSE of the bias corrected model, the bias corrected model shows significant improvement - 27\% lower RMSE for ozone - over the base forecast.
\end{abstract}

\section{Introduction}
Poor air quality and its effect on human health is an important topic as extreme air quality events are increasing and contributing to an increase in mortality \cite{roberts2021global,hou2016long}.  Both increased levels of ozone and fine particulate matter (PM\textsubscript{2.5}) have negative effects on human health, potentially causing early death in vulnerable populations \cite{bowe2019burden,giannadaki2016implementing,huangfu2020long}.  

Operational air quality (AQ) forecasting for the United States is performed by the National Oceanic and Atmospheric Administration (NOAA),the Environmental Protection Agency (EPA) and state and local agencies.  This forecasting is critical to vulnerable populations.  Changes made to the AQ forecasts can have a large implications on the population.  For this reason, any changes made to the operational AQ forecast systems requires rigorous processing, testing, and validation.  When systematic and deterministic errors are present that the currently deployed model does not address, a post-processing correction is applied to correct for these errors after the forecast model runs.  This method of post-processing correction is called bias correction.  

In this work we describe a new approach to AQ forecasting model bias correction that uses a cluster of bi-directional Long Short Term Memory (LSTM) deep learning network. The forecast-aware architecture for the LSTM uses a stacked bidirectional LSTM architecture \cite{graves2005bidirectional} and performs learning forward and background in time.  

LSTMs have shown to be useful for learning time series data due to their ability to capture long term memory \cite{greff2016lstm}. LSTMs are a type of Recurrent Neural network (RNN) that consists of a chain of cells, each containing a memory and hidden state which are passed on to the next cell in the chain. Gates within each cell allow an input value taken from an input data series to modify the hidden state, which then modifies the memory state.

\begin{figure}
\centering
\includegraphics[width=.98\columnwidth]{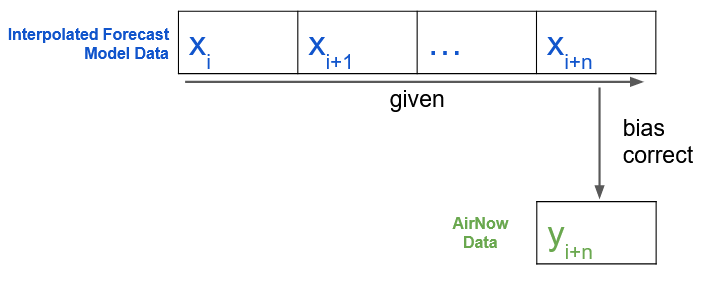}
\caption{Bias Correcting Forecast Data}
\label{fig:approach}
\end{figure}

In the described approach the LSTM setup includes a sequence of forecast data where the input corresponds to AQ forecast time intervals and the prediction is the next step of the corresponding AQ observation, depicted in Figure \ref{fig:approach}.  

Using an unsupervised clustering approach that clusters station data according to a set of features, including latitude and longitude and  urbanization, this method is able to identify regions of commonality among associated with the physical location of each AQ sensor.  Each region defines a LSTM and the training data that will be used. 

By creating these regions based on common features, each LSTM contributes to an overall improved performance rather than using one monolithic LSTM.  Our method provides a way to make the LSTM clusters more sensitive to changes by having each a region of commonality that is able to accommodate including meteorological features as part of the clustering.  When evaluating this approach, RMSE scores when compared with forecast RMSE show substantial improvement on average. 

\section{Background}
NOAA National Weather Service (NWS) provides operational (AQ) forecasting which includes ozone and fine particulate matter (PM\textsubscript{2.5}).  These forecasts are used by state and local agencies for official AQ forecasts and provide protection of life and public health.  The next generation of the NOAA AQ forecasting will include a rapid refresh forecast system coupled to the Community Multiscale Air Quality chemistry model (RRFS-CMAQ) and integrated machine learning to increase forecasting performance enabling higher resolution model predictions.  The effort described in this work is being developed for use with the currently operational AQ system and to support eventual high-resolution, AI-driven RRFS-CMAQ system for multi-day operational AQ predictions.  

\section{Bias Correction Using Traditional Methods}
Bias correction methods used for the NOAA operational AQ forecasts have traditionally been based on statistical approaches that require historical information pertaining to the NOAA weather forecasting model and observational data.  Methods such as Kalman filters are effective estimators, but they make certain assumptions such as assuming state variables are of a Gaussian distribution.  Linear processes are also assumed with respect to state transitions. In this setting Kalman filters are used to estimate prediction error. This error is subtracted from the forecast to obtain the correction.  Advanced methods that combine Kalman filters and analog methods (ANKF) \cite{djalalova2015pm2}, which is used operationally at NOAA, tend to perform well for systematically identified bias and have some adaptability built-in for updating training data automatically with the latest data. However, this method is still limited, as new types of errors need to be identified in order for it to be applied (i.e. there is less generalizability built-in).  The Kalman filter approach can over and under estimate the correction of model bias in the presence of extreme AQ events.  Machine learning methods are advantageous due to their ability to generalize and handle nonlinear problems, as they are nonparameteric.  Machine learning methods do not make the same Gaussian assumptions as the Kalman filter method. By incorporating the unsupervised clustering method with our approach, we learn a natural clustering across AQ sensors that supports a more flexible approach as the clusters can change based on the features included.

\section{Related Work}
Post processing correction of bias errors in forecast models is part of the operational workflow. Prior work in the area of bias correction has included using statistical-based methods which require a history of model output and observations \cite{huang2017improving}.  Notably, the following methods tend to be frequently referenced in terms of historical application of post-processing bias correction including:  the moving mean method \cite{zhou2017numerical}, Kalman filters \cite{djalalova2010ensemble}, analog method \cite{djalalova2015pm2}, and a combination of Kalman filters and the analog method \cite{zhou2017numerical,djalalova2015pm2,monache2014analog}.  According to \cite{huang2017improving}, these methods have been successfully applied to weather forecasts, O3 forecasts and PM\textsubscript{2.5} forecasts.  

In work by Djalalova et al. \cite{djalalova2010ensemble} a comparison was performed using these different methods.  It was found that historical analogs of hourly Kalman filtered forecasts resulted in the most accurate PM\textsubscript{2.5} forecasts with improvement of model errors when compared with raw Community Multi-scale Air Quality (CMAQ) ranging from 50 to 75\%.  Corrections were applied to the CMAQ using interpolation applied on an hourly basis.

Zhou et al. \cite{zhou2017numerical} also focused on improving PM\textsubscript{2.5} forecast models (specifically for China), they proposed a bias-correction framework which included a feature selection process, a classifier, a Kalman filter and interpolation.  

In work by Cho et al. \cite{cho2020comparative}, a number of machine learning methods were described in terms of correcting a local data assimilation and prediction system used in South Korea.

To the knowledge of the authors, this is the first time cluster-based LSTMs are used for bias correction, which could be adapted for extreme AQ events, by the use of unsupervised learning to identify regions of commonality across feature sets.

\section{The AQ Observation and Forecast Data}
Using AQ observations (which are the actual ozone/PM\textsubscript{2.5} concentrations observed at a particular time) and AQ forecast data (a set of physical variables that include ozone/PM\textsubscript{2.5} concentration), this work leverages the time-series nature of AQ forecasts.  We describe each of these dataset further.

\subsection{AirNow AQ Observation Data}
\begin{figure}
\centering
\includegraphics[width=.95\columnwidth]{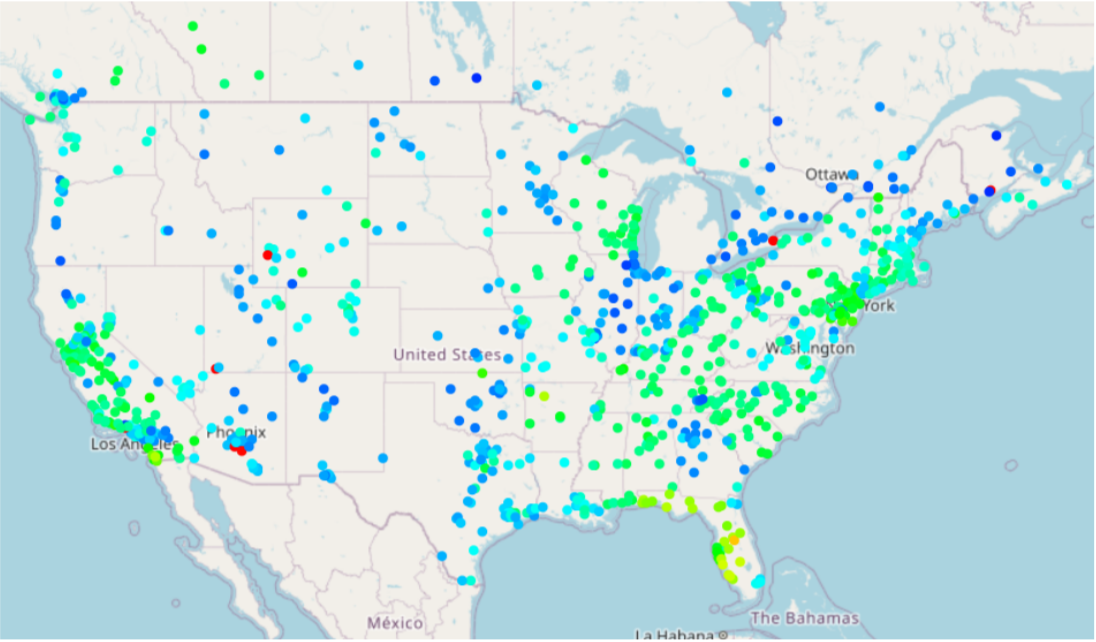}
\caption{AirNow Station Coverage. The colors indicate the amount of missing data on a scale of blue to red (blue stations are missing little data, red stations are missing large amount of data).}
\label{fig:stations}
\end{figure}

\begin{figure}
\centering
\includegraphics[width=.95\columnwidth]{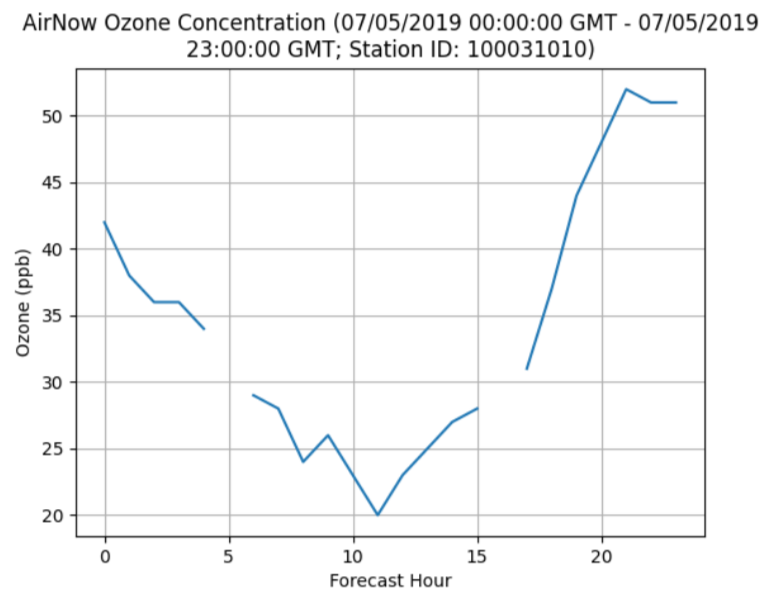}
\caption{AirNow Stations and Missing Data - Station 100031010 Located in New Castle County, Delaware.}
\label{fig:missing}
\end{figure}

Ground truth observations of ozone and PM\textsubscript{2.5} concentrations were sourced from individual AirNow \cite{schulte2020responsive} stations across the continental US and Canada.  A map of these stations is shown in Figure \ref{fig:stations}. A period of six months of observations is used, beginning in July 2019 and ending in December 2019. Nearly all stations had missing data points caused by invalid measurements at specific hours during the six month period.  We show an example of those missing data in Figure \ref{fig:missing}. If observation $y_i$ contained invalid data for hour $i$, then any set of training data that contained $y_i$ could not be used during training. Pairs containing invalid AirNow data were discarded before training began.

\subsection{NOAA AQ Forecast Data}
Forecast data was provided as a set of interpolated data derived from a gridded forecast simulation from historical NOAA AQ models to the latitudes and longitudes of AirNow observation stations. Not every station in the forecast data set corresponded to a station in the observation data set and vice versa. Only stations that appeared in both data sets were used since both forecast and observation data were required to produce training data for that station.

\subsection{AQ Data Preparation for Machine Learning}
How we use the AQ observations and forecast data is formalized as follows.  A sequence of $n$ hours of forecast data
$$\hat{x_i}=\left[x_i, x_{i+1}, x_{i+2}, ..., x_{i+n}\right]$$
(where each $x_t$ is a feature vector containing a set of physical variables for the $t$th hour of the sequence) and output a single value that is compared against the AQ observation $y_{i+n}$ at the same geographic location of the forecast corresponding to the final hour in $\hat{x_i}$ is used for training. This results in one $\hat{x_i}$ for each hour $i$ in the span of time we are trying to bias correct (except for the first $n$ hours of that time span).

Each set of forecast data was provided in individual files for each day in the training data and consisted of 48 hours of forecast values for various latitude and longitudes starting at 13:00 GMT that day. The forecast data for each hour $t$ in the forecast file is a vector $x_t$ consisting of the physical variables ozone, temperature, ground radiation, planetary boundary layer height, wind direction, wind speed, NOX, NOY, and time of day. PM\textsubscript{2.5} concentration data consists of PM\textsubscript{2.5} concentration, ground radiation, wind direction, wind speed, time of day, and RC/RN.

\subsection{Forecast Data Challenges}
Using a typical approach to data preparation for an LSTM, a single sequence of forecasts per station was generated.  To generate these sequences the final 24 hours of each forecast file were discarded so that the end of that file's forecast "day" would match up with the beginning of the next file's forecasts. Concatenating files together in this way produced a single long sequence of predictions for every hour in the training data set. Producing pairs of training data $\left(\hat{x_i},y_{i+n}\right)$ for each hour $i$ was done by taking an $n$ hour long sliding window from the beginning to the end of this concatenated series of forecast data, and matching the final hour of the window to the corresponding $y_{i+n}$ observation. This sliding window would cover all but the first $n$ hours of the training data as outputs, meaning that almost every hour in the set could be bias corrected with this method.

Many issues arose with the way the forecast data was generated using this approach. Since forecasts were generated for each day using known physical variables for that day, the beginning of the next day's forecast could be radically different from the 24 hour mark of the previous day's forecast. This often produced a large unnatural jump between two values in some $\hat{x_i}$ sequences when the sequence contained data from two different days' forecast files. In such cases, there would exist vectors $x_k$ and $x_{k+1}$ where $x_k$ would belong to the previous day's forecast file and $x_{k+1}$ would belong to the next day's forecast. This often resulted in large jumps in the values for various physical variables between the the two vectors as the forecast was recalculated using updated values for the physical variables each day. 

When we initially used this data approach to train the LSTM, it would learn from these large jumps even though they were non-physical. Limiting the LSTM to only train "within" each forecast file was also infeasible using this method since it would only be able to start bias correcting for the last $48-n$ hours. A different approach was needed to solve these problems.

\section{Forecast-Aware Bi-Directional Triad LSTM Network}
The primary obstacle using this approach to training an LSTM is that the first $n$ hours of each day of a forecast required some number of hours of input from the previous day's forecast, which could potentially include a large nonphysical jump. The only way to prevent this would be to restrict the LSTM to only take in inputs that lay within a single forecast file, but this prevented the first $48-n$ hours of forecasts for each day from being bias corrected.  Instead of using a single LSTM that required the previous 12 hours of forecast data as input, we used a novel LSTM approach which we call a forecast-aware "triad" LSTM network and includes a "pre" LSTM for the first $n$ hours, an "end" LSTM for the final $n$ hours in each 48 hour file, and a "mid" LSTM for the remaining hours.

\begin{figure}
\centering
\includegraphics[width=.95\columnwidth]{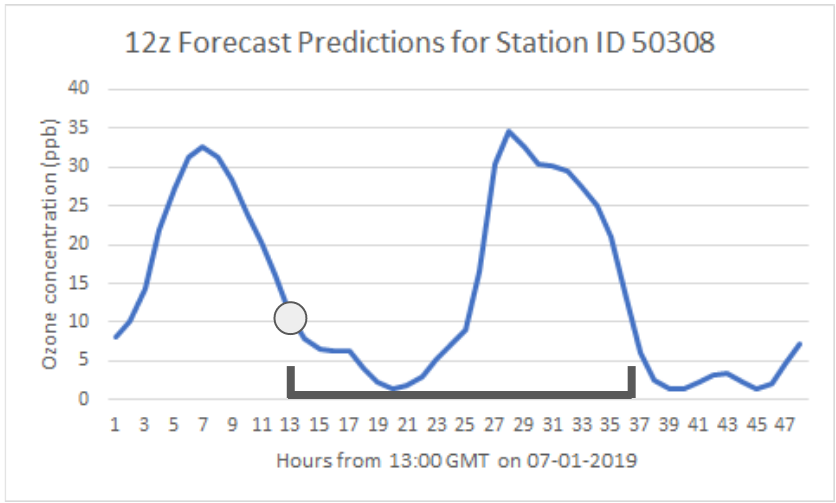}
\caption{Forecast-aware LSTM Triad "Pre"-LSTM.}
\label{fig:pre}
\end{figure}

\begin{figure}
\centering
\includegraphics[width=.95\columnwidth]{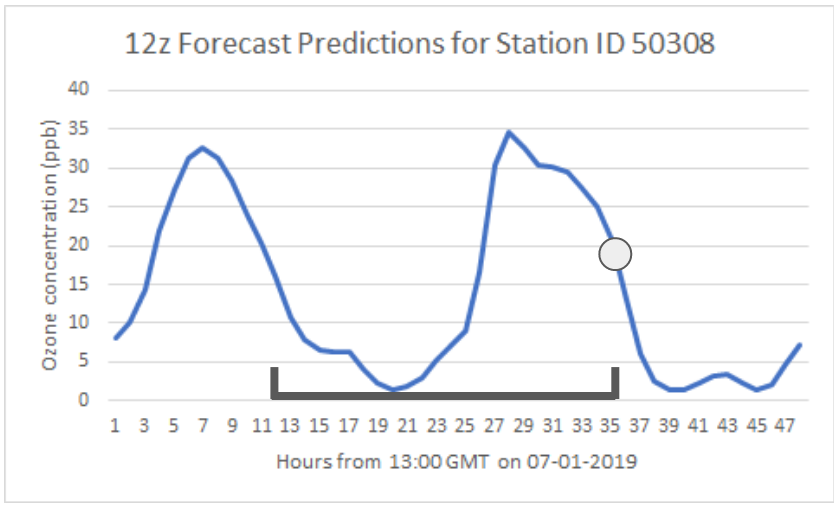}
\caption{Forecast-aware LSTM Triad "End"-LSTM}
\label{fig:end}
\end{figure}

\begin{figure}
\centering
\includegraphics[width=.95\columnwidth]{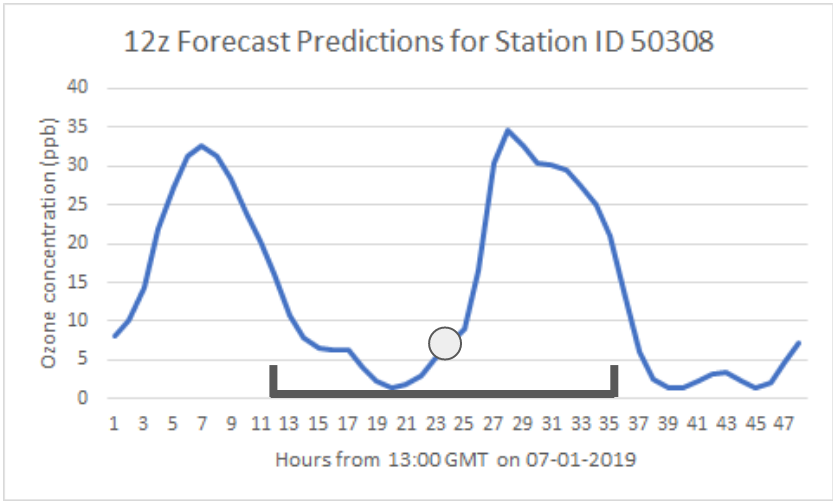}
\caption{Forecast-aware LSTM Triad "Mid"-LSTM}
\label{fig:mid}
\end{figure}

The "pre" LSTM takes in $n$ hours of data \textit{starting} at the hour we are trying to bias correct and ending $n$ hours later depicted in Figure \ref{fig:pre}. The "end" LSTM functions identically to the LSTM described above, taking in a window of $n$ hours ending with the hour we are trying to bias correct as input, depicted in Figure \ref{fig:end}. The "mid" LSTM takes in $n$ hours of forecast data, $\left\lceil n/2 \right\rceil$ before the hour we are trying to bias correct (including the hour itself) and $\left\lfloor n/2 \right\rfloor$ after, depicted in Figure \ref{fig:mid}. The sequence of forecast data in the "pre" LSTM is reversed so that the LSTM learns the sequence backwards, the bias corrects the last hour (which corresponds to first hour in the original sequence).

We experimented using the triad LSTM approach on single stations using a sliding window of $n$ hours of forecast data, but instead of moving the window over the entire set of forecast data, it is moved over the 48 hours of single day's forecast file. For each sequence
$$\hat{x_i}=\left[x_i, x_{i+1}, x_{i+2}, ..., x_{i+n}\right]$$
we produce three different outputs, one for each LSTM. The "pre" LSTM output is compared to the observation $y_{i}$, the "mid" output is compared to $y_{\left\lceil i+n/2 \right\rceil}$, and the "end" output is compared to $y_{i+n}$. This generates pairs $\left(\hat{x_i},y_{i}\right)$, $\left(\hat{x_i},y_{\left\lceil i+n/2 \right\rceil}\right)$, $\left(\hat{x_i},y_{i+n}\right)$ that are used as training data for the "pre", "mid", and "end" LSTMs. Since $\hat{x}$ stays entirely within a single day's forecast file, we avoided the issue of nonphysical jumps in the forecast data between days. When correcting a day's forecast file, we use one of the three LSTMs depending on which hour of the forecast we are correcting. 

We applied this methodology to bias correct AQ forecasts for all of the continental US and Canada (CONUS).  However, the forecast-aware triad LSTM was unable to capture the dynamical nature of the problem.  We then explored using the forecast-aware triad LSTM for different regions of the Continental United States (CONUS).  Preliminary explorations into this setup involved grouping together stations by radius and then by US state and EPA region. While some states and regions (such as Colorado) showed good results, others did not. This result motivated the use of an unsupervised method for clustering stations by features.

\section{Unsupervised Learning for Clustering AQ Stations}
K-means clustering was used to cluster stations (i.e. sensors) by  region using various features. K-means clustering \cite{hartigan1975k} is an algorithm capable of partitioning $n$ points of data into $k$ different clusters so that the sum of squared distances between each point within each cluster and the center of the cluster in minimized. More formally, given a set of points $C$ and a distance $d(x,y)$ function between two points $x,y\in C$, we seek a partition of $C$
($\left\{C_1,C_2,...,C_k \right\}$)
such that
$$\sum_{i=1}^{k}\sum_{x\in C_i}d\left(x,\mu_i\right)^2$$
is minimized where $\mu_i$ is the center of the cluster $i$ computed as
$$\mu_i=\frac{1}{\left|C_i\right|} \sum_{y\in C_i} y$$

Each region forms a set of common stations that could then be used to train the forecast-aware triad LSTM  for that region.  

Regions are clustered based on features including latitude, longitude, elevation, and urbanization. Urbanization was measured using USDA-provided rural-urban commuting area (RUCA) codes, which measure the urbanization of a region on a scale of 1 to 10. Individual stations were matched to their corresponding RUCA regions to determine their urbanization. Elevation for individual stations were determined by looking up their latitude and longitude using the USGS Elevation Point Query Service \cite{hollister2017elevatr}. 

By combining the unsupervised learning with the forecast-aware triad LSTM approach, we were able to achieve impressive results when compared with the forecast alone.  We now describe the experimental setup.

\section{Experimental Setup}
We performed ozone experiments using the forecast-aware triad LSTM with and without the regional clustering. As part of this exploration we evaluated different LSTM architectural choices, including experiments varying the dropout (from $0$ to $0.5$ in steps of $0.05$) and the number of units (from $50$ to $200$ in steps of $50$) and layers (from $1$ to $5$ in steps of $1$) for the LSTMs. A value of $0.25$ for dropout with $200$ units and $5$ layers were found to produce better performance, and were used for all future experiments. 

Model effectiveness was evaluated by running the model on the forecast of a set of testing days that were held out of the training data for each station. The resulting set of forecast and bias corrected predictions were compared against the set of actual observed concentrations for each hour and station in the test days. The Root Mean Squared Error (RMSE) of both the forecast and bias correction model was then calculated from these data sets, and the difference between the forecast and bias corrected RMSE was used to measure the performance of the bias correction model. Additional measures that were considered included RMSE for just the first 24 hours of a day, Pearson correlation \cite{pearson1909determination} across the entire day, the difference between the forecast/bias corrected maximum concentration and the observed maximum concentration for the first 15 and final 33 hours.

We also compared the forecast-aware triad LSTM approach with a Gated Recurrent Unit (GRU) approach, as recent literature suggests some performance gains could be achieved using GRUs \cite{chung2014empirical} in terms of training time, complexity of the network and reduction in training data needed to achieve the same performance.

\subsection{Defining Regions Using Unsupervised Learning}
The training data for each station in a cluster was combined together to train the triad LSTM, and two randomly selected days per month across all stations in the cluster were held out to use to evaluate the performance of the model. Various values of $k$ (specifically, $20$, and $25$) were used with the goal of finding the value that produced the greatest reduction in RMSE between the original forecast and bias corrected model. In early experiments, values of $K=7$, $10$, and $15$ were used with a different set of stations, and the reduction in RMSE was found to increase with the value of $K$ increasing, which is why $20$ and $25$ were chosen for the most recent experiments. 

Once this value of $K$ was found, it was used to perform a different set of experiments that included additional geographic features for clustering. Three experiments were performed in total: one that included both elevation and urbanization, one that included just elevation, and one that included just urbanization. Stations where elevation or urbanization could not be determined were excluded from the clustering and experiments.  Each feature was normalized and given equal weight to every other feature during clustering.

The six month period of forecast data also included large gaps of missing data in the middle of December. Just like with invalid observation data, all pairs of training data that included forecast data from this period were discarded prior to training. Furthermore, no training days were taken from the December data due to the large gaps.

For each cluster, the bias correction model trained on all training data for each station in the cluster, then ran on the held out set (July 13th and 23rd, August 4th and 18th, September 8th and 20th, October 1st and 30th, and November 9th and 22nd 2019) for each station to evaluate its performance.

\section{Results and Analysis}
We describe the results for experiments that were conducted for the forecast-aware single station LSTMs and also for the cluster-generated triad LSTMs, given different cluster sizes and variables.  We provide metrics in terms of the overall average RMSE and also look at individual stations.  For all of the experiments that follow we used the six months of data as described in the data sections.

\subsection{Forecast-Aware Triad LSTM Single Station Results}
In the forecast-aware single station experiments, the bias correction model was trained using data from a single station. The forecast for two randomly chosen days per month was held out from training so they could be used to evaluate the model's performance. We also performed similar experiments for PM\textsubscript{2.5}.  For those experiments, the triad LSTM makes large improvement over the forecast trend, however for some stations the triad LSTM does not reproduce high frequency variability in the AirNow data well.

Single station models showed promising initial results, but were limited by the fact that in order to bias correct all stations in the CONUS, three LSTMs would need to be trained per station. This quickly became intractable given the number of AirNow stations in operation. Furthermore, training a single set of LSTMs per station limited us to just that station's worth of data. Since each LSTM is trained on only its station's data, they tend to be overfit and are unable to generalize to other stations.  

\subsection{Cluster-Generated Forecast-Aware Triad LSTM Results}

\begin{figure}
\centering
\includegraphics[width=.99\columnwidth]{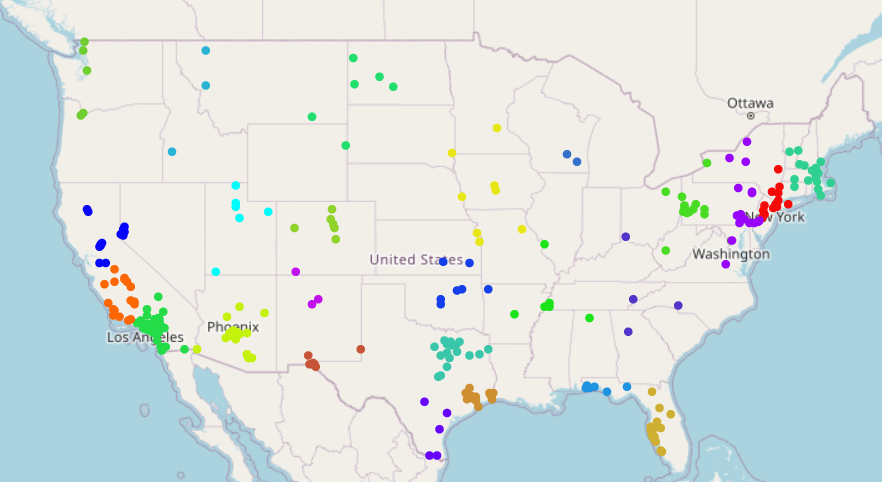}
\caption{Clustering Stations Based on Latitude and Longitude for K=25.}
\label{fig:k25}
\end{figure}

\begin{figure}
\centering
\includegraphics[width=.99\columnwidth]{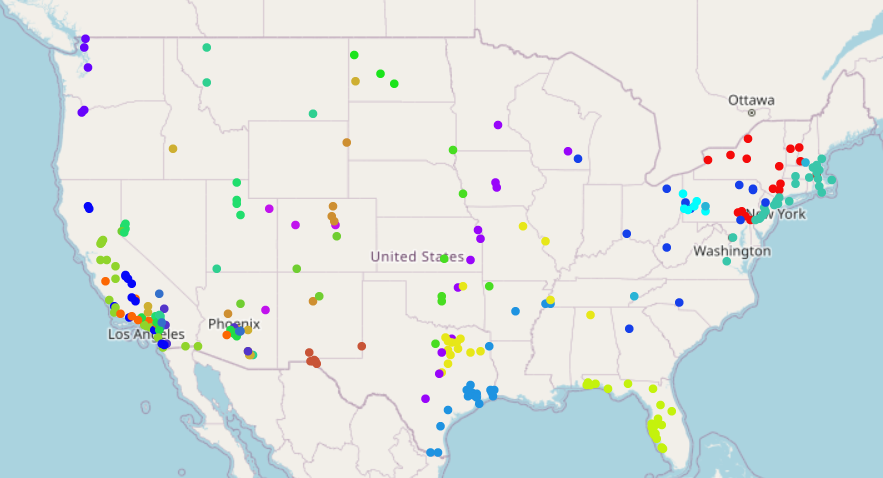}
\caption{Clustering Stations Based on Latitude, Longitude, and Elevation for K=25.}
\label{fig:k25elevation}
\end{figure}

\begin{figure}
\centering
\includegraphics[width=.99\columnwidth]{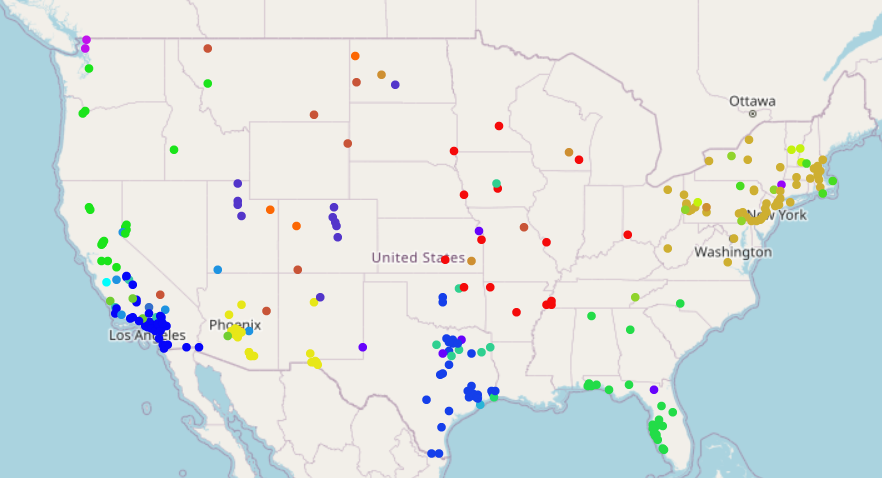}
\caption{Clustering Stations Based on Latitude, Longitude, and Urbanization for K=25.}
\label{fig:k25urbanization}
\end{figure}

We visually show the clustering outcomes for K=25 for the different clustering variables (latitude/longitude, latitude/longitude/elevation,  latitude/longitude/urbanization, and latitude/longitude/elevation/urbanization) in Figures \ref{fig:k25}, \ref{fig:k25elevation} and \ref{fig:k25urbanization} where different colored dots represent different clusters.

Results for varying $K$ are summarized in Table \ref{tab:kmeans}. The $k=25$ clusters showed the greatest reduction ($2.8435$ ppb of ozone) in RMSE between the forecast and bias corrected model. The reduction increases with the size of $k$, which is expected since the model needs to generalize for smaller and smaller sets of stations. The reduction for having a uniquely trained model for each station is 3.604 ppb, but this reduction requires training 271 different models, none of which are trained to generalize to other stations.

\begin{table}
\centering
\begin{tabular}{ |l||l|l|l|l|  }
\hline
RMSE (ozone)&K=20&K=25&Single\\
\hline
\hline
Forecast&11.9253	&11.9253	&11.9253\\
\hline
Bias Corrected&8.7981	&8.7859 &8.321\\
\hline
Reduction&3.1272	&3.1394	&3.604\\
\hline
\end{tabular}
\caption{Summary of results for the cluster-based triad LSTMs for various values of K and the single station triad LSTM results across all testing days.}
\label{tab:kmeans}
\end{table}

Results for using different clustering variables are summarized in Table \ref{tab:variables}. The combination of latitude and longitude, and elevation (excluding urbanization) provided the largest reduction (3.2726 ppb of ozone).

\begin{table}[H]
\centering
\begin{tabular}{ |l||l|l|l|l|  }
\hline
RMSE (ozone)&L&L/E&L/U&L/E/U\\
\hline
\hline
Forecast&11.9253	&11.9253	&11.9253   &11.9253\\
\hline
Bias Corrected&8.7859	&8.6527 &9.0157   &8.7780\\
\hline
Reduction&3.1394	&3.2726	&2.9096&   3.1473\\
\hline
\end{tabular}
\caption{Summary of results for $K=25$ using different clustering features. Each column title indicates the feature used - "L" represents latitude and longitude, "E" represents elevation, and "U" represents urbanization.}
\label{tab:variables}
\end{table}

\begin{figure}[H]
\centering
\includegraphics[width=.99\columnwidth]{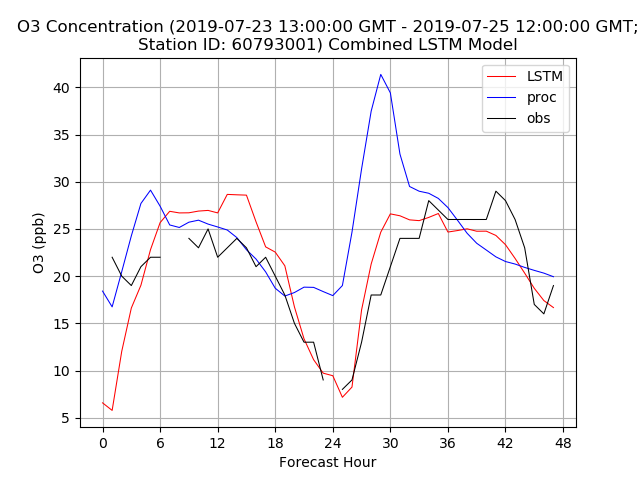}
\caption{Ozone LSTM Results for K=25 with latitude and longitude as clustering variables - Station 60793001 Located in Morro Bay, California. Proc represents the forecast mode, obs represents the AirNow ground truth, and LSTM represents the result of the triad LSTMs.}
\label{fig:morrobay_25}
\end{figure}

\begin{figure}
\centering
\includegraphics[width=.99\columnwidth]{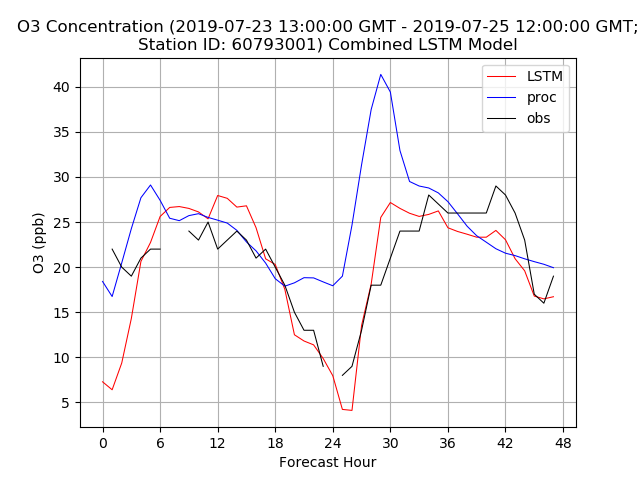}
\caption{Ozone LSTM Results for Single Station - Station 60793001 Located in Morro Bay, California. Proc represents the forecast mode, obs represents the AirNow ground truth, and LSTM represents the result of the triad LSTMs.}
\label{fig:morrobay_single}
\end{figure}

\begin{figure}[t]
\centering
\includegraphics[width=.99\columnwidth]{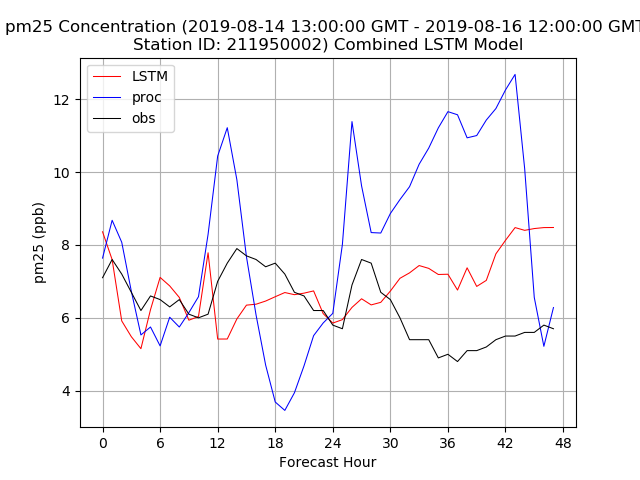}
\caption{PM\textsubscript{2.5} K=25 cluster results for station 211950002 (located in Pikeville, Kentucky). Proc represents the forecast mode, obs represents the AirNow ground truth, and LSTM represents the result of the triad LSTMs.}
\label{fig:pm25}
\end{figure}

We also highlight results from the cluster-based triad LSTM in Figure \ref{fig:morrobay_25} for Station 60793001 using $K=25$ with latitude and longitude as clustering variables in comparison with the Single Station results in Figure \ref{fig:morrobay_single}. Note that the bias corrected predictions for the clustered-based LSTMs are similar to the predictions for the Single Station LSTM despite not being specifically trained on just those stations. Our initial PM\textsubscript{2.5} results showed a reduction in RMSE of 1.945 ppb over the forecast model. We show the result for one such station in Figure \ref{fig:pm25}.

Finally, an experiment was run where each stacked bidirectional LSTM in the forecast-aware triad LSTM model was replaced with a stacked bidirectional GRU. In this triad GRU model, the training time was reduced by 10 hours (from 70 to 60) but the reduction in RMSE was also reduced by approximately 0.07 ppb of ozone (from 2.843 to 2.773).  These results show a benefit in reducing the training time.  Future experiments will explore comparing the LSTM with GRU by reducing the overall training data.


\section{Conclusions and Future Work}
In this paper we describe a novel clustering-based forecast-aware triad LSTMs.  Operational requirements can challenge how a deep learning method could be included in AQ forecasting. Given these requirements and limitations in the data, along with the prospect of new extreme AQ events, we have proposed a way to address these challenges.  The dynamic nature of AQ suggests that using the clustering method may enable the LSTM to adjust for extreme AQ events by allowing for additional meteorological features to be included as part of the clustering.  We show that the cluster-based forecast-aware LSTMs perform significantly better than the base forecast bias correction on a 2019 six month dataset.  Future experiments will explore training on the year 2019 and testing on the year 2020, which is rich in extreme AQ events due to wildfire activity.

Additional experiments will also be performed which compare the LSTM approach with existing bias correction methods deployed at NOAA. We are also experimenting with larger test sets across days spanning different seasons and additional features for the clustering method.

The challenge with K-means clustering is that one needs to define the number of clusters a-priori. We obtained $K$ by experimenting with different values for $K$.  However, we would like to learn $K$ automatically.  A way to address this issue would be to implement a Hierarchical Dirichlet Process method \cite{teh2006hierarchical} that could automatically learn the appropriate number of clusters.

\bibliography{bias_correct_lstm}
\end{document}